\documentclass[
  reprint,
  amsmath,amssymb,
  aps,
  pre,
  floatfix,
]{revtex4-2}

\usepackage{graphicx}
\usepackage{dcolumn}
\usepackage{bm}
\usepackage{subfig}
\usepackage[colorlinks=true,citecolor=blue,linkcolor=blue,urlcolor=blue]{hyperref}
\usepackage[mathlines]{lineno}

\begin{document}

\title{Sequential Learning and Catastrophic Forgetting in Differentiable Resistor Networks}

\author{Maniru Ibrahim}
\email{maniru.ibrahim@ul.ie}
\affiliation{Mathematics Applications Consortium for Science and Industry (MACSI), Department of Mathematics and Statistics, University of Limerick, Limerick, Ireland}

\begin{abstract}
Differentiable physical networks provide a simple setting in which learning can
be studied through the interaction between trainable parameters and physical
equilibrium constraints. We investigate sequential learning in differentiable
resistor networks governed by Kirchhoff's laws. Although individual
input--output mappings can be learned by gradient-based adjustment of edge
conductances, sequential training on conflicting tasks produces catastrophic
forgetting. We show that forgetting is controlled by task conflict and by the
degree of adaptation to the new task. Uniform anchoring and normalised
gradient-weighted anchoring reduce forgetting only by increasing the final
loss on the new task, giving a clear forgetting--adaptation trade-off. We also
show that forgetting is associated with localised conductance changes on
high-current edges, giving a physical interpretation as reconfiguration of
dominant transport pathways. Broader random-task ensembles show that the
strongest forgetting occurs when the second task reverses the output ordering
imposed by the first task. Finally, comparisons across Erdős--Rényi,
small-world, scale-free, and random-geometric graph ensembles show that
topology changes the forgetting--adaptation balance. These results position
differentiable resistor networks as compact, physically interpretable testbeds
for studying continual learning in tunable matter.
\end{abstract}

\maketitle

\section{Introduction}

Recent years have seen growing interest in \emph{physical neural networks} and, more broadly, in learning systems that compute directly through physical dynamics rather than purely digital numerical operations \citep{wright2022deep,momeni2023backpropfree,momeni2024training,stern2023learningwithoutneurons}. Optical, electronic, mechanical, and other analog substrates can transform inputs through their intrinsic dynamics, while trainable internal parameters allow those substrates to be adapted to desired tasks. This line of work is attractive both practically, because of its potential energy and latency advantages, and conceptually, because it provides a route to studying learning as a physical process rather than solely as an abstract algorithmic one.

Among such systems, resistor networks offer a particularly transparent setting. Their physical state is determined by Kirchhoff's laws, their trainable parameters have a direct interpretation as conductances, and their input--output map is naturally expressed through a graph Laplacian. In this sense, differentiable resistor networks occupy a useful middle ground: they are simple enough to analyse mechanistically, yet rich enough to display nontrivial learning behaviour \citep{stern2021supervised,stern2025become,guzman2025microscopic}. They therefore provide a natural platform for studying how learning, memory, and interference arise in tunable physical media.

A central challenge in machine learning is \emph{continual learning}, in which a system must acquire tasks sequentially without erasing previously learned behaviour. Standard gradient-based training is well known to suffer from \emph{catastrophic forgetting}, whereby optimisation for a new task degrades performance on earlier tasks \citep{french1999catastrophic,kirkpatrick2017overcoming,zenke2017si,serra2018hat,delange2021survey}. This issue is especially relevant for hardware-aware and neuromorphic systems, where online adaptation, limited memory, and local constraints make full rehearsal or repeated retraining difficult \citep{davies2018loihi,furber2016large,indiveri2015memory}. Despite this, forgetting has been studied far less in differentiable physical networks than in software neural networks.

At the same time, the physical-learning literature has begun to move beyond single-task trainability toward questions of adaptation, robustness, and structure formation in tunable networks \citep{stern2021supervised,dillavou2024machine,stern2024power,stern2025become,guzman2025microscopic}. Most directly relevant is the recent preprint of \citet{chatterjee2025remembrance}, which studies catastrophic forgetting in tunable resistor networks and shows that thresholded local updates can preserve memory of multiple sequential tasks by spatially confining training updates. Our paper is therefore not framed as introducing another forgetting-mitigation rule for tunable resistor networks. Instead, we focus on a different and complementary contribution: a differentiable, gradient-based, graph-Laplacian study of forgetting that emphasises mechanism. In particular, we analyse forgetting through the geometry of task gradients, the localisation of conductance updates, and the preferential rewiring of high-current pathways.

Concretely, we make five contributions. First, we formulate a minimal
sequential-learning benchmark in differentiable resistor networks and show that
conflicting tasks produce catastrophic forgetting under ordinary gradient-based
training. Second, we show that forgetting is governed by a
forgetting--adaptation trade-off: stronger adaptation to the new task generally
causes stronger degradation of the old task. Third, we compare uniform anchoring
with normalised gradient-weighted anchoring and show that both move the system
along this trade-off. Fourth, we broaden task conflict beyond a single reversed
task by sampling random continuous target pairs, showing that forgetting is
strongest when the second task reverses the output ordering imposed by the
first. Fifth, we test robustness across graph topology, input--output graph
distance, network size, and training budget. The resulting picture is that
catastrophic forgetting in this class of physical networks is not merely an
abstract optimisation failure, but a physically interpretable and
topology-dependent reconfiguration of conductive pathways.

\section{Related Work}

\subsection{Continual learning and catastrophic forgetting}

Catastrophic forgetting has been recognised for decades as a central limitation of sequential learning in connectionist models \citep{french1999catastrophic}. Modern continual-learning methods address this problem through several broad strategies: regularisation-based approaches constrain parameter updates, replay-based methods reuse previous data or synthetic surrogates, and architectural approaches reduce interference through masking, modularisation, or task-specific subspaces \citep{kirkpatrick2017overcoming,zenke2017si,li2018lwf,serra2018hat,delange2021survey}. Among these, regularisation-based methods are the most relevant to the present work because they formalise forgetting as parameter drift away from previously useful configurations. Our anchor penalty belongs to this family, though in a deliberately simplified form suited to the resistor-network setting.

\subsection{Neuromorphic and hardware-constrained continual learning}

A second relevant literature concerns learning in hardware-constrained and neuromorphic systems. In such settings, continual learning is shaped not only by algorithmic considerations but also by locality, memory, energy consumption, and device-level constraints \citep{davies2018loihi,furber2016large,indiveri2015memory}. These systems motivate continual-learning strategies that are lightweight, online, and physically implementable. From this perspective, differentiable physical networks are especially interesting: they inherit many of the same constraints, but they also expose a direct relationship between physical transport, trainable parameters, and task interference.

\subsection{Physical neural networks and learning in physical systems}

Physical neural networks and self-learning physical systems have recently emerged as a broad research area spanning optics, electronics, mechanics, and dynamical systems \citep{wright2022deep,momeni2023backpropfree,momeni2024training,stern2023learningwithoutneurons}. Work in this area has addressed both trainability and implementation, including equilibrium-based learning, physical reservoir computing, and local learning in analog substrates \citep{scellier2017equilibrium,jaeger2004harnessing,tanaka2019recent,stepney2024tutorial}. In parallel, \citet{stern2021supervised} formulated supervised learning rules for physical networks more generally, while recent experiments have demonstrated processor-free or decentralized learning in nonlinear analog networks and self-learning circuits \citep{dillavou2024machine,stern2024power}. Most of this literature, however, has focused on how physical systems can be trained at all, rather than on how they remember or forget tasks under sequential learning.

\subsection{Adaptive and tunable physical networks}

Within physical learning, tunable resistor and flow-like networks are particularly relevant because they combine simple physical laws with highly interpretable trainable structure. The work of \citet{stern2025become} and \citet{guzman2025microscopic} shows that learned physical networks acquire measurable structural signatures tied to function, and that trained solutions leave microscopic imprints in tunable media. The recent preprint of \citet{chatterjee2025remembrance} takes an important step further by directly studying sequential learning in tunable resistor networks and showing that thresholded local updates can reduce catastrophic forgetting by spatially separating task-specific tuned regions. Our paper differs in both method and emphasis. Methodologically, we study a differentiable resistor-network model trained by automatic differentiation rather than a thresholded local learning rule. Conceptually, we focus less on proposing a new local mitigation mechanism and more on explaining forgetting itself through gradient conflict, update localisation, current-weighted parameter change, and task recovery.

Related work has also examined interference and memory in physical learning
systems more broadly. Falk et al.~\cite{falk2023learning} studied learning multiple
functions in mechanical networks and analysed how task compatibility affects
whether functions can be embedded simultaneously. Dillavou et
al.~\cite{dillavou2025understanding} investigated contradictory tasks in physical
learning systems and the conditions under which learning one task interferes
with another. The present work differs in that we focus specifically on
differentiable resistor networks trained by gradient descent through a
graph-Laplacian equilibrium solve, and we analyse forgetting through
forgetting--adaptation trade-offs, high-current pathway reconfiguration, and
graph-topology dependence.
\section{Differentiable Resistor Network Model}

We consider an electrical network with $N$ nodes and $E$ resistive edges. Each edge $e=(i,j)$ carries a positive conductance $w_{ij}>0$, and each node has an electrical potential $v_i$. The network is thus represented as a weighted graph whose edge weights are trainable physical parameters.

For nodes whose voltages are not externally prescribed, the equilibrium state is determined by Kirchhoff's current balance:
\begin{equation}
\sum_j w_{ij}(v_i-v_j)=0.
\end{equation}
Writing \(v\in\mathbb{R}^N\) for the vector of node voltages and \(L(w)\) for
the weighted graph Laplacian, the imposed input voltages are treated as
Dirichlet boundary conditions. Partition the nodes into prescribed boundary
nodes \(B\) and free nodes \(I\). After imposing the boundary voltages \(v_B\),
Kirchhoff's equations for the free nodes become
\begin{equation}
L_{II}(w)v_I=-L_{IB}(w)v_B .
\label{eq:reduced_laplacian}
\end{equation}
The output is read from the component of the equilibrium voltage vector
corresponding to the designated output node.

Two nodes are designated as input nodes, and their voltages are fixed externally. A separate node is designated as the output node, and its equilibrium voltage is interpreted as the network prediction. The resulting input--output map is therefore not specified directly; instead, it emerges implicitly from the conductance-dependent equilibrium response of the physical network.

To guarantee positivity of conductances during optimisation, each conductance is parameterised by an unconstrained variable $\theta_k$ through a softplus transformation,
\begin{equation}
w_k = \log\!\left(1+e^{\theta_k}\right)+\epsilon,
\end{equation}
where $\epsilon>0$ is a small numerical floor. This reparameterisation allows unconstrained optimisation in $\theta$-space while preserving physical admissibility of the conductances.

Given an input pair $x$, the model prediction is the voltage at the output node after solving the equilibrium system. Learning is performed by minimising a squared-error loss between this output voltage and a prescribed target. Because the equilibrium solve is differentiable, gradients of the loss with respect to $\theta$ can be computed by automatic differentiation through the linear system. The model therefore belongs to the class of differentiable physical networks, but unlike many black-box implementations, both the operator $L(w)$ and the learned parameters retain a clear electrical and graph-theoretic interpretation.

\section{Sequential Task Training}

To probe continual-learning behaviour in the simplest possible setting, we define two binary input--output tasks on the same pair of input nodes. Task A is
\begin{align}
(1,0) &\rightarrow 1,\\
(0,1) &\rightarrow 0,
\end{align}
whereas Task B reverses these targets:
\begin{align}
(1,0) &\rightarrow 0,\\
(0,1) &\rightarrow 1.
\end{align}
These tasks are intentionally conflicting: they impose opposite desired responses on the same input patterns. This makes them a useful minimal benchmark for isolating interference and forgetting.

Sequential learning proceeds in two stages. The network is first trained on Task A from an initial parameter vector $\theta_0$, yielding a learned configuration $\theta_A$. Training is then continued on Task B starting from $\theta_A$. Forgetting is quantified by the increase in Task A loss induced by the second stage:
\begin{equation}
F = L_A^{\mathrm{after}} - L_A^{\mathrm{before}},
\end{equation}
where $L_A^{\mathrm{before}}$ is the Task A loss immediately after the first training stage and $L_A^{\mathrm{after}}$ is the Task A loss after Task B training. Large positive values of $F$ indicate strong forgetting.

This two-task construction is deliberately minimal. It is not intended as a realistic benchmark for large-scale continual learning, but as a controlled setting in which the geometry of gradient interference and the physical structure of conductance reconfiguration can be analysed transparently.

\section{Experimental Results}

The experiments are organised around three questions. First, how does task
conflict control forgetting? Second, how does regularisation move the system
along a forgetting--adaptation trade-off? Third, how do physical structure and
graph architecture shape this trade-off?

Unless otherwise stated, the baseline Erdős--Rényi experiments use \(N=40\),
edge probability \(p=0.15\), learning rate \(0.1\), and 300 gradient steps on
each task. The topology experiments use \(N=80\), approximately matched mean
degree, and 500 gradient steps on each task. Reported error bars denote one
standard deviation across graph realisations. For topology comparisons,
forgetting is interpreted jointly with final Task B loss, since reduced
forgetting may reflect incomplete learning of the second task rather than
improved memory retention alone.
\subsection{Baseline sequential learning}

\begin{figure}[t]
\centering
\includegraphics[width=\columnwidth]{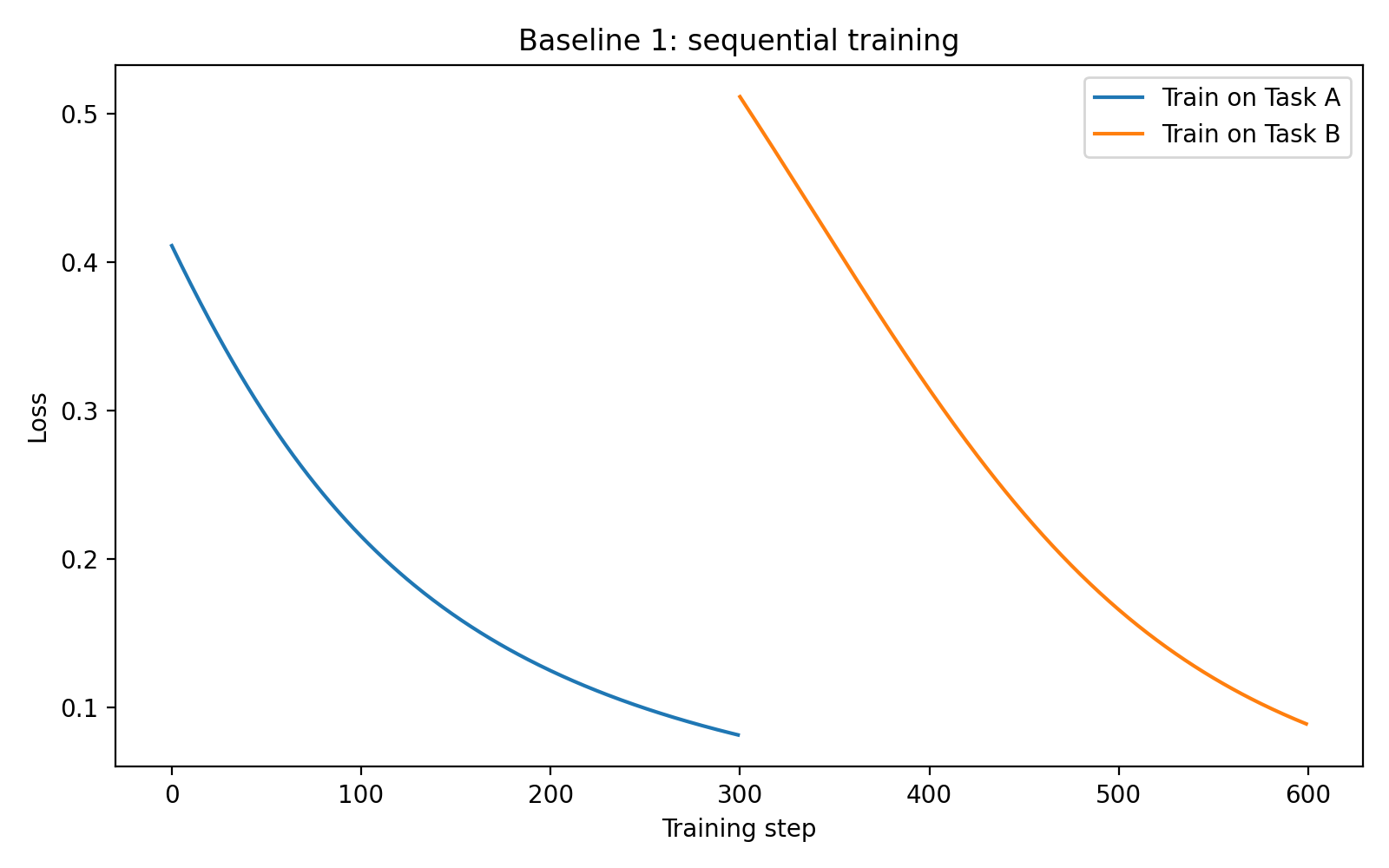}
\caption{Sequential training behaviour in the resistor network for a representative run. The model is first trained on Task A and then on Task B. While both tasks can be learned individually, Task A performance deteriorates substantially after Task B training, demonstrating catastrophic forgetting.}
\label{fig:baseline_training}
\end{figure}

We first examine the unregularised sequential training procedure. Figure~\ref{fig:baseline_training} shows the loss curves for a representative run.

During Task A training, the loss decreases from approximately $0.41$ to $0.08$, demonstrating that the resistor network can successfully learn Task A. When training is subsequently switched to Task B, the Task B loss decreases from approximately $0.51$ to $0.09$, indicating that the network can also adapt to the new task.

However, this adaptation causes severe degradation of previously learned behaviour. After Task B training, the Task A loss increases from $0.081$ to $0.494$, yielding a forgetting measure of $F = 0.412$. Thus, the parameter updates required to solve Task B effectively overwrite the conductance configuration that encoded Task A.

This experiment establishes catastrophic forgetting as a clear baseline phenomenon in the differentiable resistor network.

\begin{table}[t]
\caption{Baseline sequential training results for a representative run.}
\label{tab:baseline_single}
\begin{ruledtabular}
\begin{tabular}{lc}
Metric & Value \\
\hline
Task A loss after Task A training & 0.081 \\
Task A loss after Task B training & 0.494 \\
Task B loss after Task B training & 0.088 \\
Forgetting $F$ & 0.412 \\
\end{tabular}
\end{ruledtabular}
\end{table}

\subsection{Task conflict and gradient-overlap transition}

The reversed Task A--Task B pair used above is the maximally conflicting
endpoint of a broader family of task pairs. To see how gradient conflict changes
as the second task moves from cooperative to contradictory, we constructed a
one-parameter family of Task B variants. Task A was fixed as
\[
(1,0)\rightarrow 1, \qquad (0,1)\rightarrow 0,
\]
while the second task was defined by
\[
(1,0)\rightarrow \alpha, \qquad (0,1)\rightarrow 1-\alpha ,
\]
with \(\alpha\in\{1,0.75,0.5,0.25,0\}\). Thus \(\alpha=1\)
corresponds to an identical task, whereas \(\alpha=0\) gives the fully reversed
task used in the baseline experiments.

For each value of \(\alpha\), we evaluated the gradient cosine similarity
between Task A and \(B_\alpha\) at several points along the Task A training
trajectory. If \(\theta_t\) denotes the parameter vector after \(t\) Task A
training steps, we compute
\begin{equation}
\Omega_\alpha(t)
=
\frac{
\nabla_\theta L_A(\theta_t)\cdot
\nabla_\theta L_{B_\alpha}(\theta_t)
}{
\|\nabla_\theta L_A(\theta_t)\|\,
\|\nabla_\theta L_{B_\alpha}(\theta_t)\|
}.
\label{eq:alpha_overlap}
\end{equation}
We also trained sequentially on \(A\to B_\alpha\) and measured the resulting
forgetting.

\begin{figure}[t]
\centering
\includegraphics[width=\columnwidth]{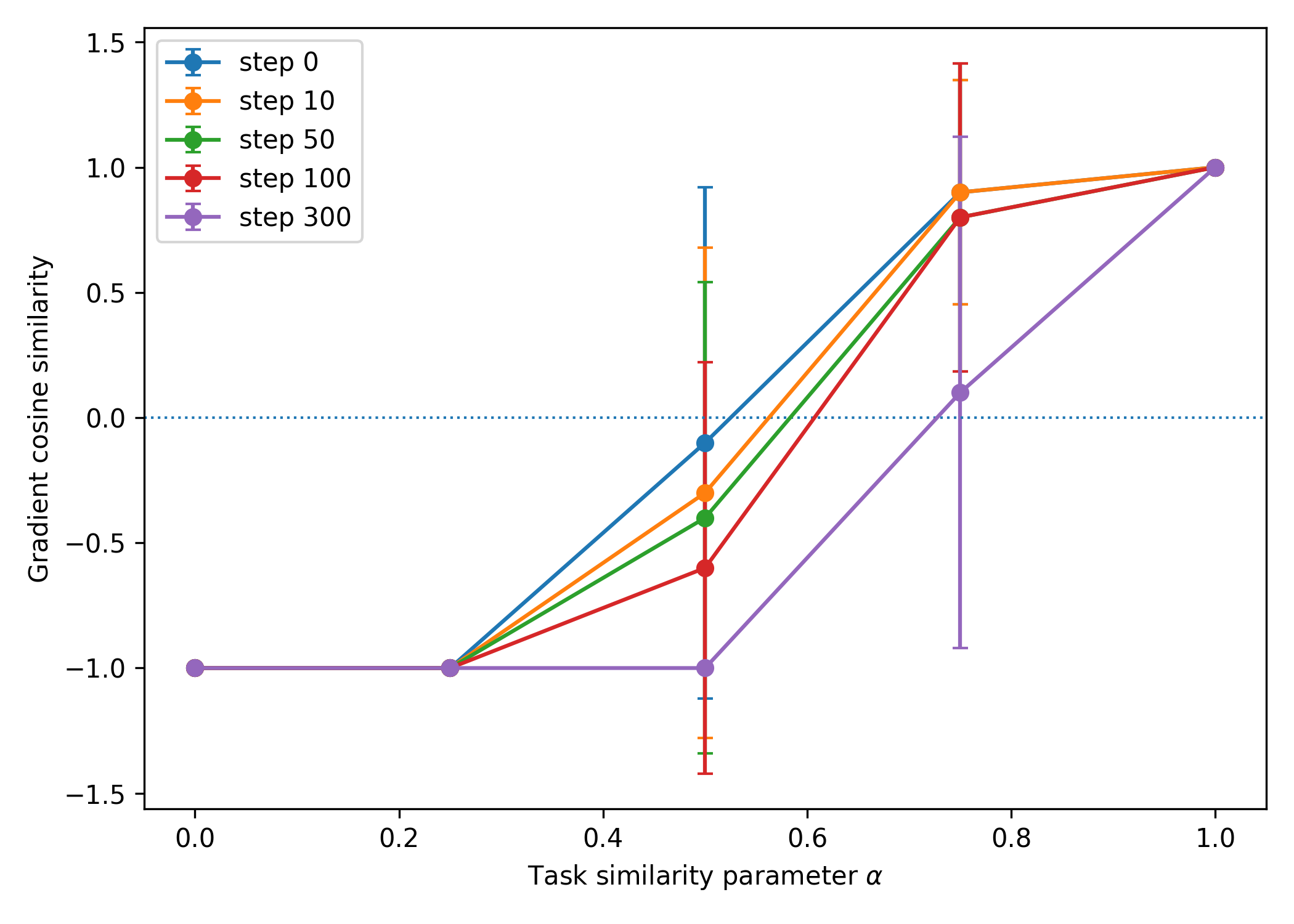}
\caption{
Gradient cosine similarity as a function of the task-similarity parameter
\(\alpha\), evaluated at different checkpoints along Task A training. The
endpoints are robust: \(\alpha=1\) gives aligned gradients, whereas
\(\alpha=0\) gives antiparallel gradients. Intermediate values show larger
variability, consistent with near-compromise tasks and smaller gradient norms.
}
\label{fig:gradient_overlap_alpha}
\end{figure}

\begin{figure}[t]
\centering
\includegraphics[width=\columnwidth]{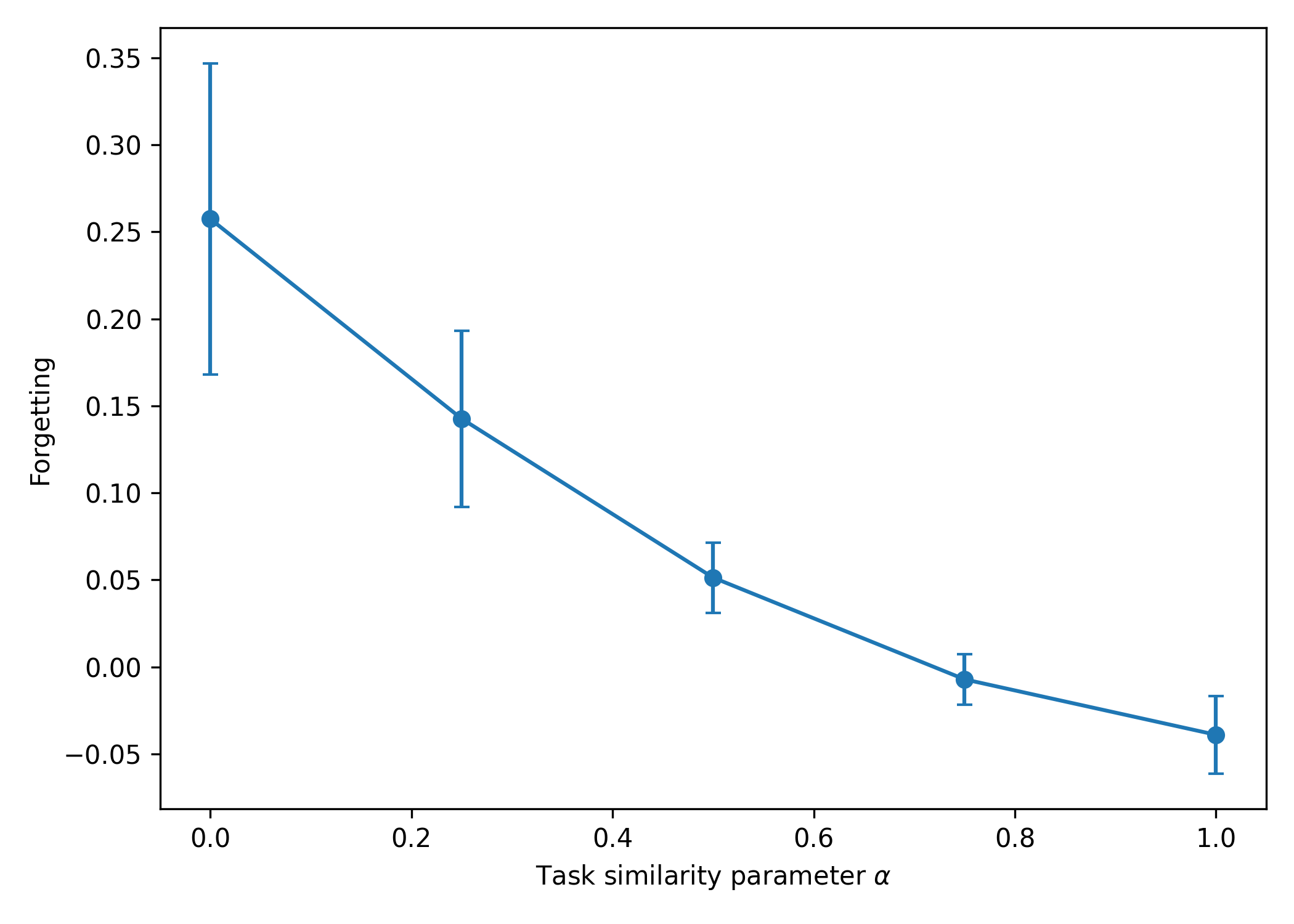}
\caption{
Forgetting as a function of the task-similarity parameter \(\alpha\). Forgetting
increases as the second task changes from identical to Task A
\((\alpha=1)\) toward the fully reversed task \((\alpha=0)\).
}
\label{fig:forgetting_alpha}
\end{figure}

Figures~\ref{fig:gradient_overlap_alpha} and~\ref{fig:forgetting_alpha}
show that the endpoints are robust. When \(\alpha=1\), the two tasks are
identical, the gradients are aligned, and forgetting is absent or slightly
negative because additional training continues to improve Task A. When
\(\alpha=0\), the tasks impose opposite targets on the same inputs, the
gradients are antiparallel, and forgetting is largest. Intermediate values of
\(\alpha\) interpolate between these regimes. The case \(\alpha=0.5\) acts as a
compromise task and produces much smaller forgetting than the fully reversed
task.

The cosine similarities show larger variability for intermediate \(\alpha\),
partly because one of the task gradients can be small near compromise
solutions. Thus, gradient overlap should be interpreted as a local diagnostic
of task conflict rather than as a complete theory of forgetting. The main
conclusion is that as the second task moves from cooperative to antagonistic
relative to Task A, the gradient directions become increasingly opposed and
forgetting increases.

\subsection{Random task ensemble and cooperative versus contradictory tasks}

The one-parameter \(B_\alpha\) family varies task conflict along a single line
in target space. To test whether the same mechanism holds for a broader class
of tasks, we sampled a random ensemble of second tasks with continuous targets
\[
(1,0)\rightarrow y_1,\qquad (0,1)\rightarrow y_2,
\]
where \(y_1,y_2\in[0,1]\). Task A was kept fixed as
\[
(1,0)\rightarrow 1,\qquad (0,1)\rightarrow 0 .
\]
For computational tractability, we used 10 graph realisations and 20 random
second tasks per graph, supplemented by the five deterministic tasks used in
the \(B_\alpha\) sweep.

In this two-input setting, a natural scalar measure of task alignment is the
Task-B target contrast
\begin{equation}
c_B=y_1-y_2 .
\label{eq:target_contrast}
\end{equation}
Since Task A has contrast \(c_A=1\), positive \(c_B\) corresponds to a task
with the same output ordering as Task A, \(c_B\simeq 0\) corresponds to a
near-neutral or compromise task, and negative \(c_B\) corresponds to a task
that reverses the Task-A output ordering.

\begin{figure}[t]
\centering
\includegraphics[width=\columnwidth]{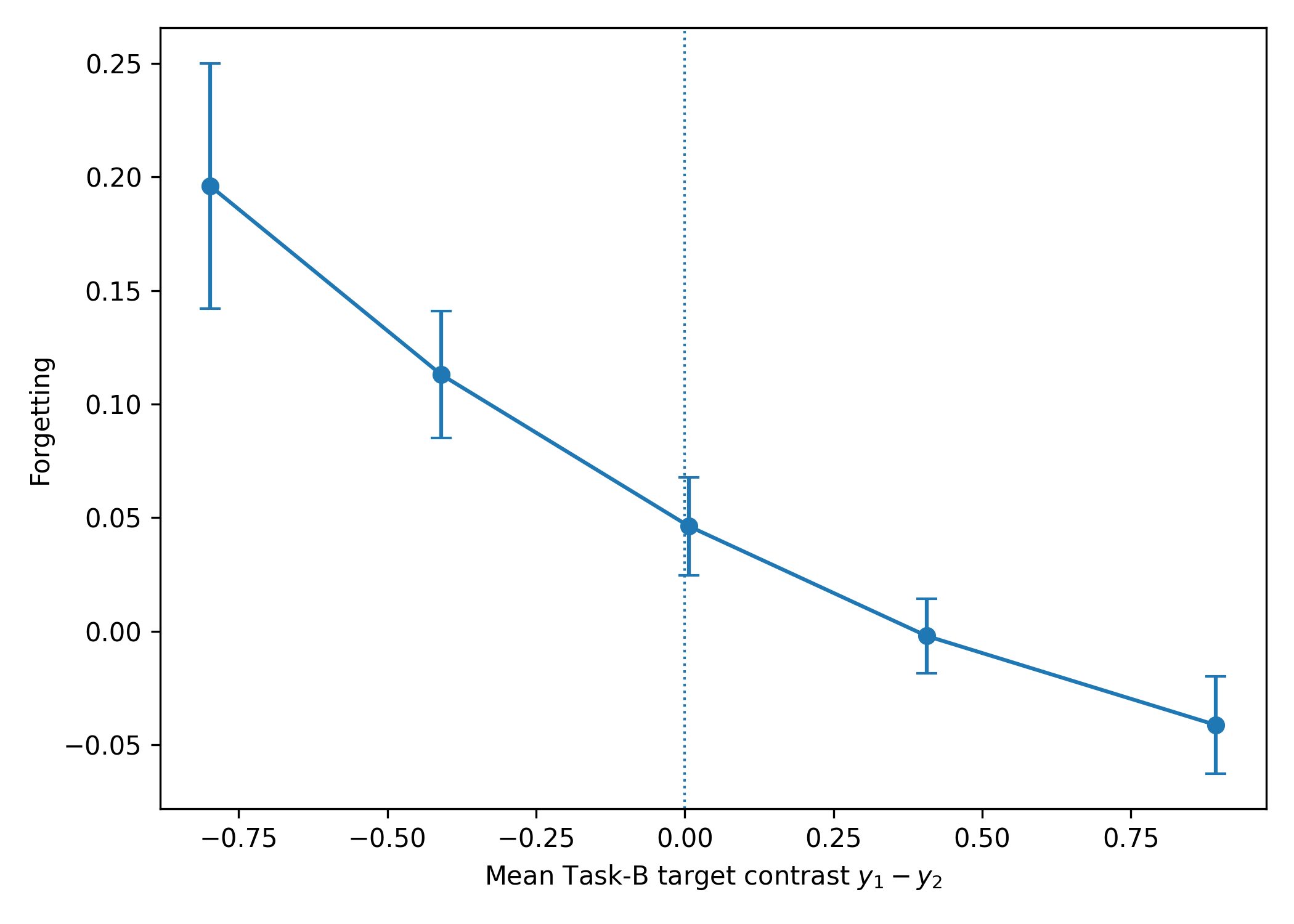}
\caption{
Forgetting as a function of the Task-B target contrast \(c_B=y_1-y_2\) for a
random ensemble of second tasks. Positive contrast corresponds to the same
output ordering as Task A, near-zero contrast corresponds to compromise tasks,
and negative contrast corresponds to reversed or contradictory tasks. Forgetting
increases strongly as the target contrast becomes negative, showing that
contradictory tasks are those that reverse the output ordering imposed by the
first task.
}
\label{fig:random_task_contrast}
\end{figure}

Figure~\ref{fig:random_task_contrast} shows forgetting as a function of
\(c_B\). The dependence is strong and monotone. Strongly negative contrasts
produce the largest forgetting, near-neutral tasks produce moderate forgetting,
and positive-contrast tasks produce little or no forgetting. Quantitatively,
the correlation between forgetting and target contrast is
\(-0.935\), while the correlation between forgetting and the mean squared
difference between Task A and Task B targets is \(0.945\). Thus, in this
minimal setting, contradictory tasks are precisely those that impose an
opposite output ordering on the same input patterns.

The initial gradient overlap also predicts forgetting, but less strongly than
the target contrast itself, with correlation \(-0.691\). This suggests that
gradient conflict is one mechanism by which target-level contradiction is
expressed in parameter space, while the target contrast provides the simpler
task-level descriptor.
\subsection{Regularisation and the forgetting--adaptation trade-off}

To characterise the stability--plasticity trade-off more systematically, we
compared uniform anchoring with a gradient-weighted anchoring baseline. The
uniform anchor penalty is
\begin{equation}
\mathcal{L}_{\mathrm{uni}}(\theta)
=
\mathcal{L}_B(\theta)
+
\lambda
\frac{1}{E}
\sum_{k=1}^{E}
(\theta_k-\theta_{A,k})^2 ,
\label{eq:uniform_anchor}
\end{equation}
where \(E\) is the number of trainable conductances.

For the gradient-weighted method, we use
\begin{equation}
\mathcal{L}_{\mathrm{gw}}(\theta)
=
\mathcal{L}_B(\theta)
+
\lambda
\frac{1}{E}
\sum_{k=1}^{E}
\widehat F_k
(\theta_k-\theta_{A,k})^2 ,
\label{eq:gradient_weighted_anchor}
\end{equation}
where \(\widehat F_k\) is a normalised importance weight. The unnormalised
importance weights are estimated from squared Task A gradients,
\begin{equation}
F_k =
\frac{1}{M}
\sum_{m=1}^{M}
\left[
\frac{\partial \ell_A^{(m)}(\theta)}
{\partial \theta_k}
\right]^2_{\theta=\theta_A},
\label{eq:importance_weights}
\end{equation}
where \(\ell_A^{(m)}\) is the loss for the \(m\)-th Task A example. We then
normalise by the mean importance,
\begin{equation}
\widehat F_k =
\frac{F_k}{E^{-1}\sum_{\ell=1}^{E}F_\ell+\varepsilon_F}.
\label{eq:normalised_importance}
\end{equation}
This normalisation separates the structure of the importance profile from its
overall numerical scale. The resulting penalty is EWC-style, but we refer to it
as gradient-weighted anchoring to emphasise that the weights are empirical
squared-gradient importances rather than a full Fisher estimate. In the present
two-example task, the per-example squared-gradient profile and the squared
gradient of the mean Task A loss were numerically indistinguishable up to
round-off.

\begin{figure}[t]
\centering
\includegraphics[width=\columnwidth]{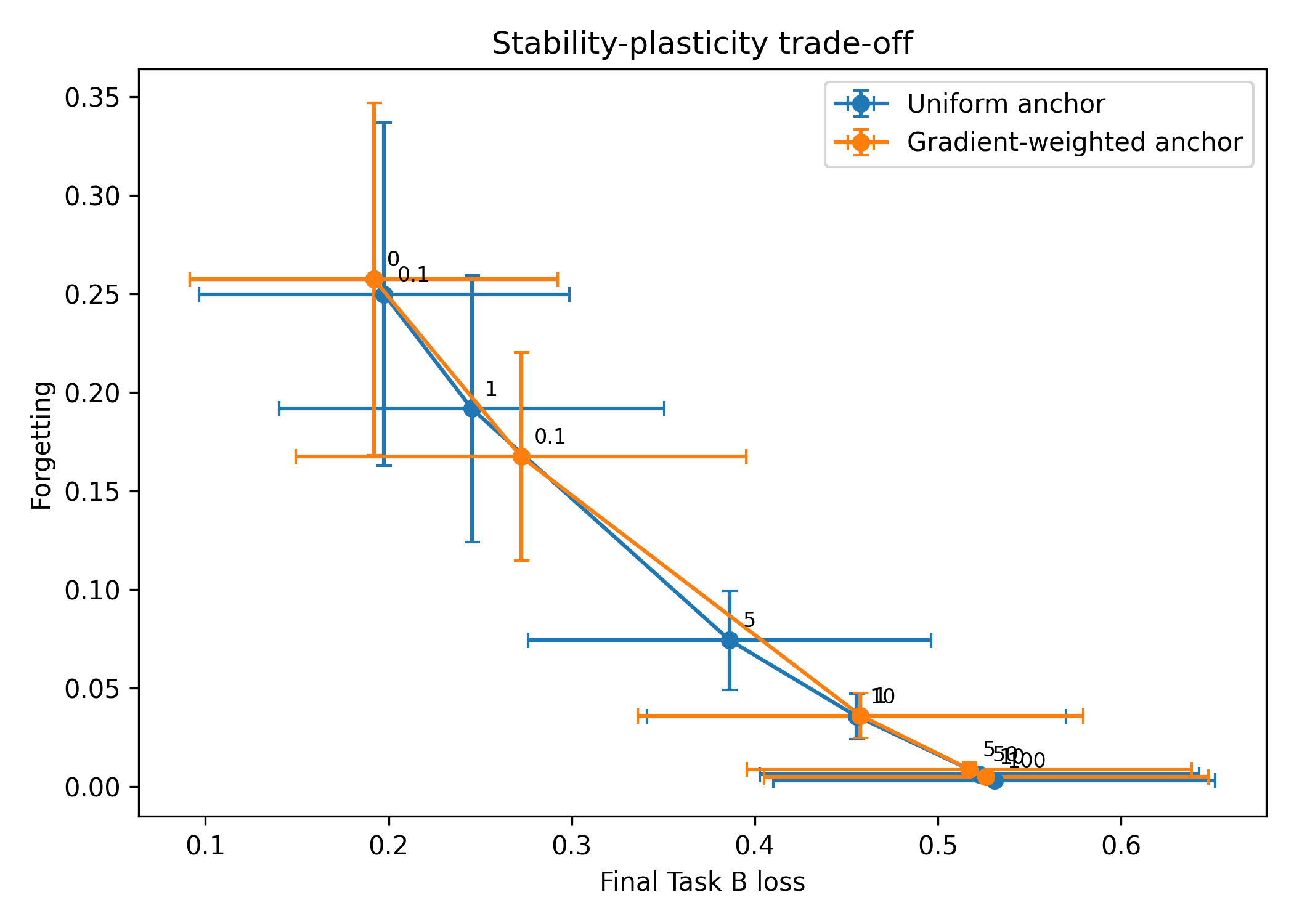}
\caption{
Forgetting--adaptation trade-off for uniform and gradient-weighted anchoring.
Each point shows the mean over 20 random graph realisations, with error bars
denoting one standard deviation. Increasing the regularisation strength reduces
forgetting but increases the final Task B loss. Gradient-weighted anchoring
produces stronger protection against forgetting at comparable \(\lambda\), but
also more strongly suppresses adaptation to Task B.
}
\label{fig:regularisation_tradeoff}
\end{figure}

Table~\ref{tab:regularisation_tradeoff} and
Fig.~\ref{fig:regularisation_tradeoff} summarise the results across 20 graph
realisations. Both regularisation methods reduce forgetting as \(\lambda\)
increases, but they do so by increasing the final Task B loss. Thus,
regularisation does not preserve Task A for free; it moves the solution along a
forgetting--adaptation trade-off. Gradient-weighted anchoring is more aggressive:
at \(\lambda=1\), it reduces forgetting to \(0.036\pm0.012\), compared with
\(0.192\pm0.068\) for the uniform anchor, but it also gives substantially worse
Task B loss.

\begin{table}[t]
\caption{
Forgetting--adaptation trade-off for uniform and gradient-weighted anchoring.
Values are mean \(\pm\) standard deviation over 20 random graph realisations.
Uniform denotes uniform anchoring and GW denotes gradient-weighted anchoring.
The gradient weights are normalised to have unit mean.
}
\label{tab:regularisation_tradeoff}
\begin{ruledtabular}
\begin{tabular}{lccc}
Method & \(\lambda\) & Forgetting & Task B loss \\
\hline
Baseline & 0
& \(0.257 \pm 0.089\)
& \(0.192 \pm 0.100\) \\
Uniform & 0.1
& \(0.250 \pm 0.087\)
& \(0.198 \pm 0.101\) \\
Uniform  & 1
& \(0.192 \pm 0.068\)
& \(0.245 \pm 0.105\) \\
Uniform  & 5
& \(0.074 \pm 0.025\)
& \(0.386 \pm 0.110\) \\
Uniform  & 10
& \(0.036 \pm 0.012\)
& \(0.456 \pm 0.114\) \\
GW & 0.1
& \(0.167 \pm 0.053\)
& \(0.272 \pm 0.123\) \\
GW & 1
& \(0.036 \pm 0.012\)
& \(0.458 \pm 0.122\) \\
GW & 5
& \(0.009 \pm 0.003\)
& \(0.517 \pm 0.121\) \\
GW & 10
& \(0.005 \pm 0.002\)
& \(0.526 \pm 0.121\) \\
\end{tabular}
\end{ruledtabular}
\end{table}

\subsection{Localisation of conductance updates}

To investigate how sequential learning modifies the network, we analysed the
change in conductance parameters during Task B training. Let \(\theta_A\) be
the parameter configuration after Task A training and \(\theta_B\) the
configuration after subsequent Task B training. For each edge,
\[
|\Delta\theta_e|=|\theta_{B,e}-\theta_{A,e}|.
\]

\begin{figure}[t]
\centering
\includegraphics[width=\columnwidth]{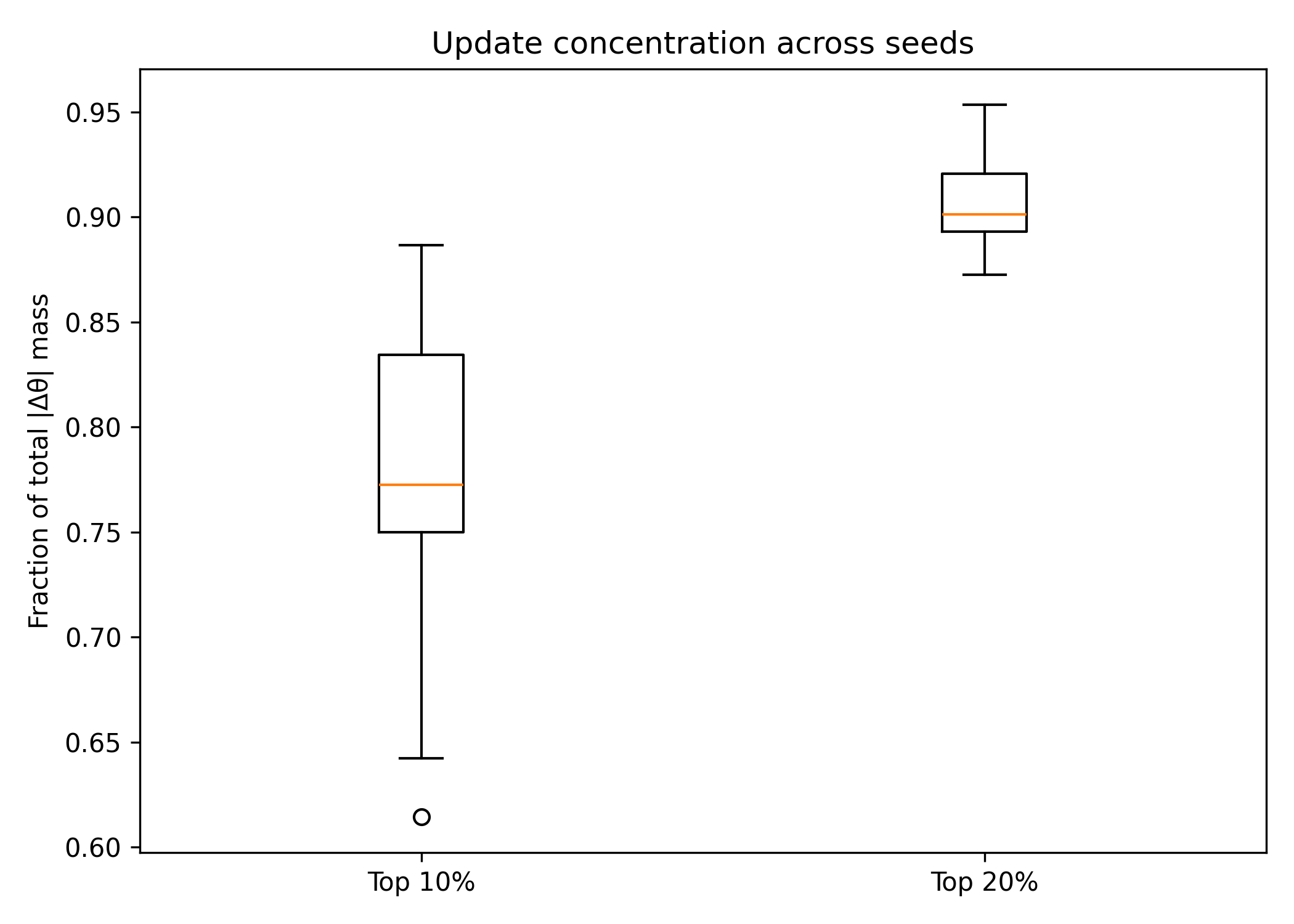}
\caption{Distribution across seeds of the fraction of total update magnitude contained in the largest edges. The top $10\%$ of edges carry most of the update mass, and the top $20\%$ account for almost all of it, showing strongly localised network rewiring.}
\label{fig:update_concentration}
\end{figure}

The update distribution is strongly heterogeneous. To quantify this
localisation, we measured the fraction of total update magnitude contained in
the largest edges across ten graph realisations. The top \(10\%\) of edges
account for \(0.772\pm0.089\) of the total update mass, while the top \(20\%\)
account for \(0.906\pm0.024\). The distribution is shown in
Fig.~\ref{fig:update_concentration}. Sequential learning therefore does not
produce uniform parameter drift. Instead, adaptation to the new task is
concentrated on a relatively small subset of conductances.

\subsection{Edge current predicts conductance updates}

To identify which edges are modified during sequential learning, we compared
the Task-B parameter update with the current carried by each edge under the
Task-A solution. For edge \((i,j)\), the current magnitude is
\[
I_{ij}=w_{ij}|v_i-v_j|,
\]
where \(w_{ij}\) and \(v_i,v_j\) are evaluated after Task A training. We then
compared \(I_{ij}\) with
\[
|\Delta\theta_{ij}|=|\theta^{(B)}_{ij}-\theta^{(A)}_{ij}|.
\]

\begin{figure}[t]
\centering
\includegraphics[width=\columnwidth]{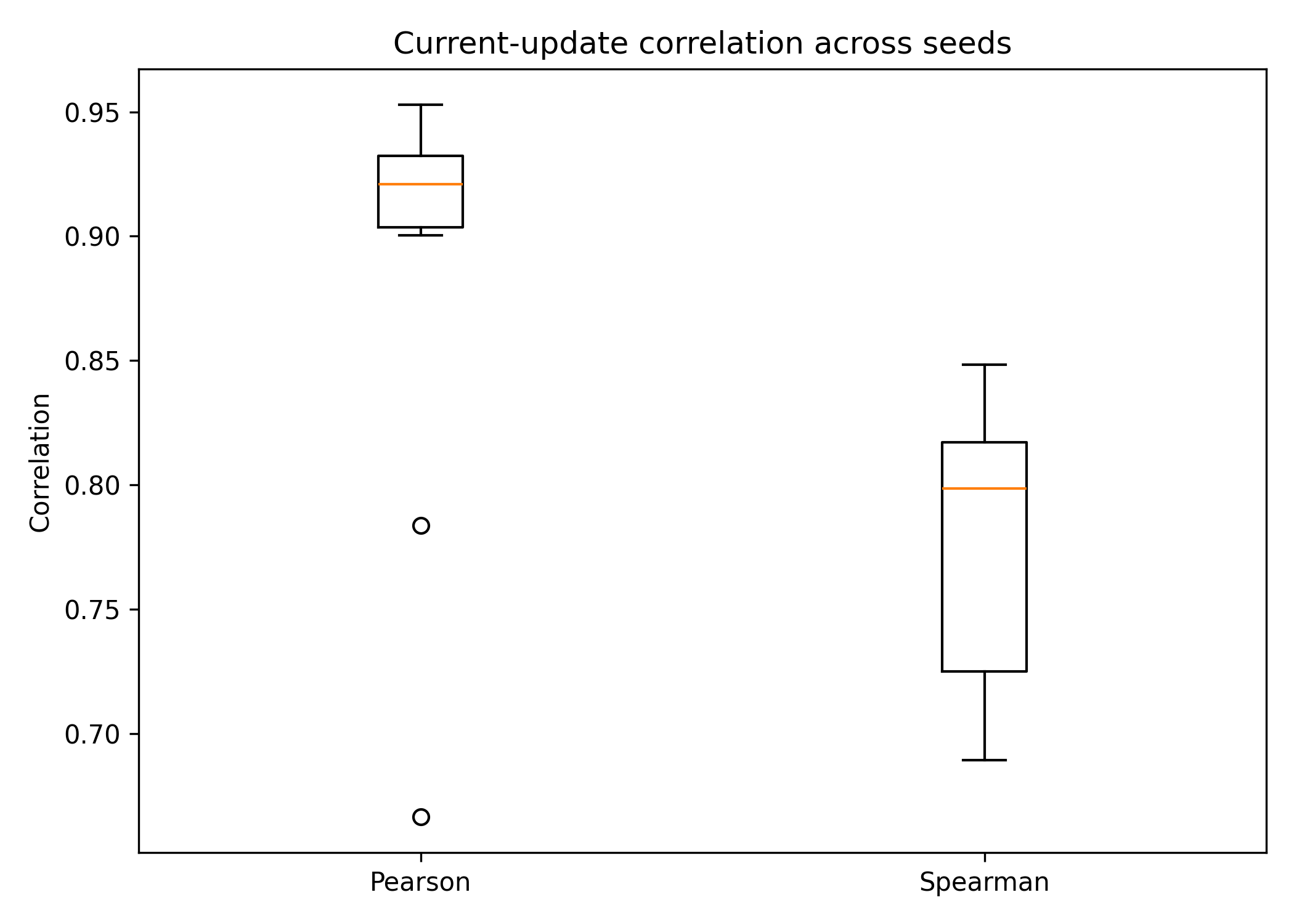}
\caption{Distribution of current--update correlations across random seeds. Both Pearson and Spearman correlations show a strong positive relationship between edge current magnitude under Task A and the magnitude of conductance updates during Task B training.}
\label{fig:current_update_correlation}
\end{figure}

Across ten graph realisations, the Pearson correlation between edge current
and update magnitude is \(0.887\pm0.091\), while the Spearman rank correlation
is \(0.776\pm0.057\), as shown in Fig.~\ref{fig:current_update_correlation}.
Thus, learning preferentially modifies edges that already carry substantial
electrical current. Forgetting therefore has a physical interpretation:
learning a new task rewires the high-current transport backbone of the network.

\subsection{Task recovery under repeated training}

To examine whether previously learned behaviour can be recovered after forgetting, we considered the task sequence
\[
A \rightarrow B \rightarrow A.
\]
The network was first trained on Task A, then on Task B, and finally retrained on Task A starting from the post-Task-B parameters.

\begin{figure}[t]
\centering
\includegraphics[width=\columnwidth]{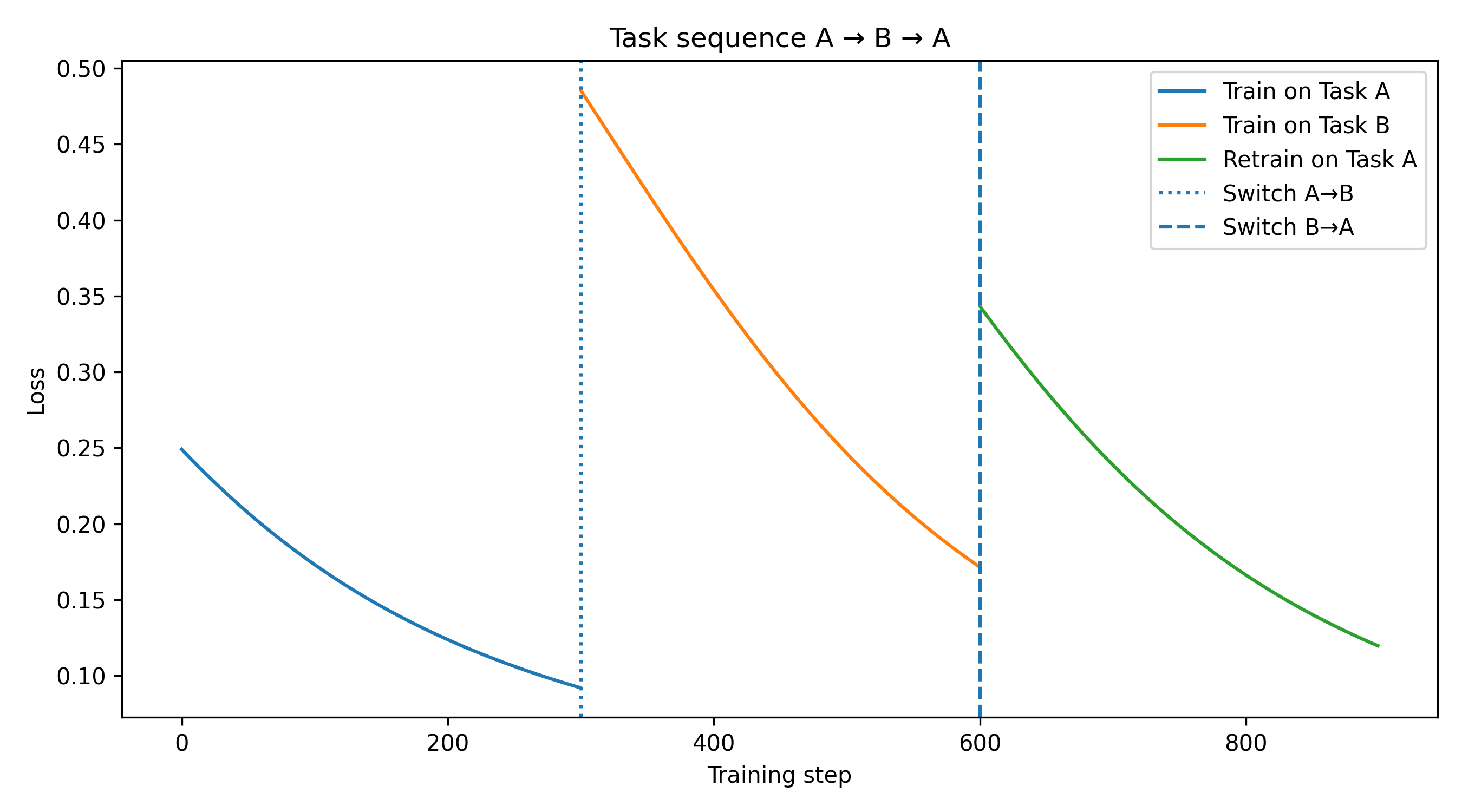}
\caption{Training losses for the task sequence $A \rightarrow B \rightarrow A$. Task A is first learned, then degraded by training on Task B, and finally partially recovered by retraining on Task A.}
\label{fig:task_recovery_curves}
\end{figure}

Figure~\ref{fig:task_recovery_curves} shows the corresponding loss trajectories. During the first training phase, the Task A loss decreases steadily to a low value. Training on Task B then increases the Task A loss substantially, confirming that the parameters learned for Task A are disrupted by subsequent optimisation on the conflicting task. When the network is retrained on Task A, the loss decreases again, indicating clear recovery of the original task. Quantitatively, the Task A loss decreases to $0.092$ after the initial Task A training phase, increases to $0.343$ following Task B training, and then decreases again to $0.119$ after retraining on Task A. Thus, retraining substantially restores Task A performance, although it does not fully recover the minimum loss reached during the first Task A phase within the same optimisation budget.

This result suggests that forgetting in the resistor network is not caused by irreversible destruction of learning capacity. Instead, sequential learning redirects the conductance configuration toward the currently trained task, and previously successful behaviour can be re-established through further optimisation. At the same time, the incomplete recovery observed here indicates hysteresis or path dependence in the parameter trajectory.

\subsection{Effect of network topology}

To test whether the observed forgetting behaviour is specific to the
Erdős--Rényi networks used in the baseline experiments, we repeated the
sequential \(A\to B\) protocol across several graph ensembles with comparable
mean degree. We considered Erdős--Rényi graphs, Watts--Strogatz small-world
graphs, Barabási--Albert scale-free graphs, and random geometric graphs. For
each graph realisation we recorded the forgetting measure, final Task B loss,
edge count, mean shortest-path length, clustering coefficient, and degree
variance. For the random-geometric ensemble, runs in which Task A was not
learned were excluded from the forgetting statistics using the criterion
\(L_A^{\mathrm{before}}<0.25\), since forgetting is not meaningful when the
initial task has not been acquired.

\begin{figure}[t]
\centering
\includegraphics[width=\columnwidth]{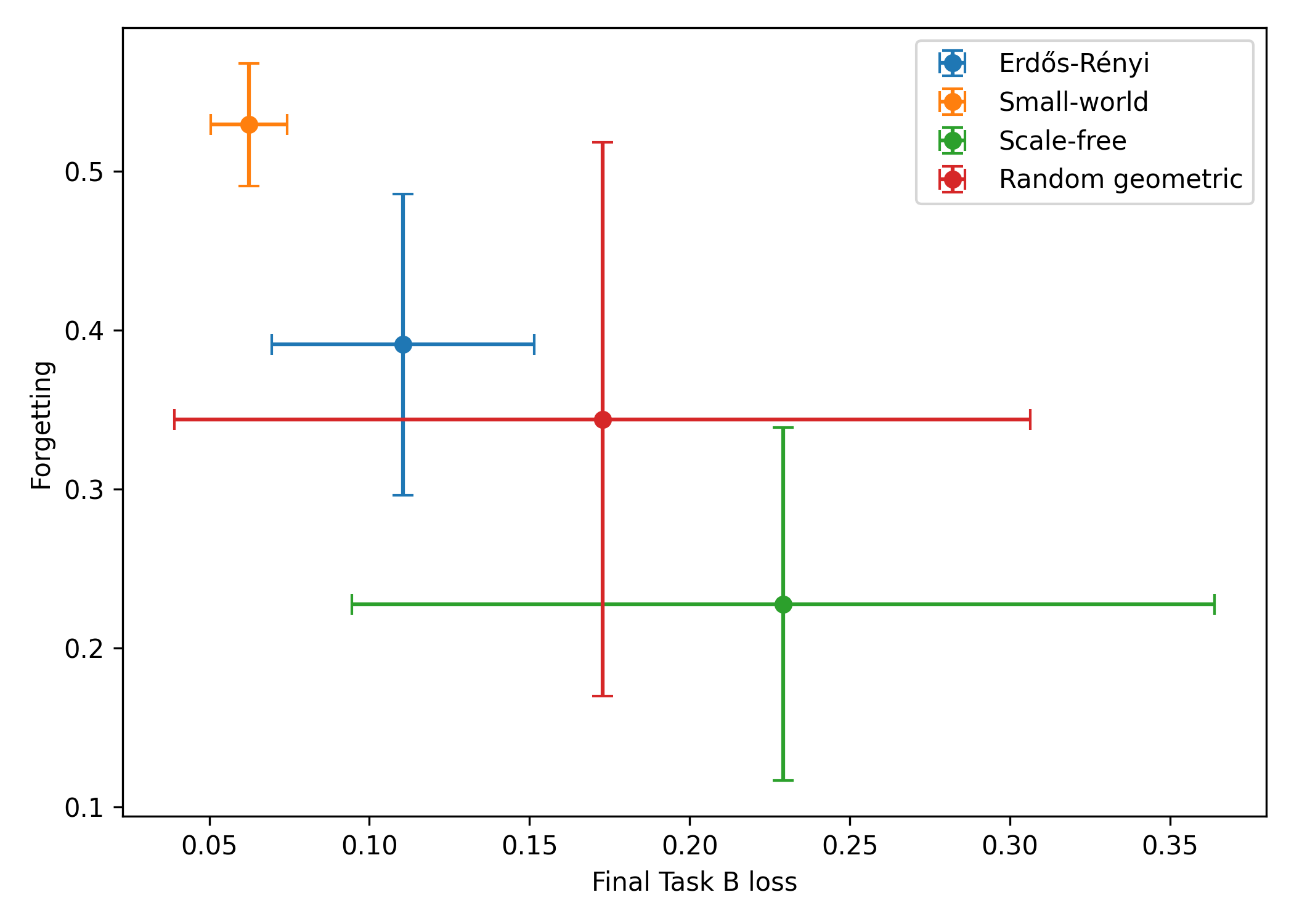}
\caption{
Effect of graph topology on the forgetting--adaptation trade-off. Each point
shows the mean over graph realisations, with error bars denoting one standard
deviation. Small-world graphs show the strongest adaptation to Task B but also
the largest forgetting, whereas scale-free graphs show lower forgetting but
weaker Task B adaptation. Random-geometric graphs show intermediate behaviour
with larger variability. Reduced forgetting should therefore be interpreted
jointly with new-task performance.
}
\label{fig:topology_tradeoff}
\end{figure}

The results are summarised in Table~\ref{tab:topology} and
Fig.~\ref{fig:topology_tradeoff}. Topology has a clear effect on the
forgetting--adaptation balance. Small-world graphs give the strongest
adaptation to Task B, as indicated by the lowest final Task B loss, but also
exhibit the largest forgetting. Scale-free graphs show the opposite tendency:
forgetting is lower on average, but Task B adaptation is weaker and more
variable. Random-geometric graphs, after controlling for edge density, show
intermediate behaviour with larger variability across realisations.

These results show that forgetting cannot be interpreted independently of
new-task learning. A topology with lower forgetting may preserve the first task
more effectively, but it may also simply learn the second task less completely.
The relevant object is therefore the joint forgetting--adaptation trade-off,
rather than forgetting alone.

For \(N=80\), we used \(p=0.08\) for Erdős--Rényi graphs,
\(k=6\) and rewiring probability \(0.2\) for Watts--Strogatz graphs,
\(m=3\) for Barabási--Albert graphs, and radius \(r=0.17\) for random
geometric graphs. The random-geometric radius was chosen empirically so that the retained graphs
had an edge count comparable to the other ensembles.

In Table~\ref{tab:topology}, \(F\) denotes forgetting, \(L_B\) denotes the
final Task B loss, \(E\) is the number of edges, and \(C\) is the clustering
coefficient. The graph classes are Erdős--Rényi (ER), Watts--Strogatz
small-world (SW), Barabási--Albert scale-free (BA), and random geometric (RG).
For the RG ensemble, we retained 17 of 20 realisations satisfying
\(L_A^{\mathrm{before}}<0.25\).
\begin{table}[t]
\caption{
Topology dependence of forgetting and Task B adaptation. Values are
mean \(\pm\) standard deviation across graph realisations.
}
\label{tab:topology}
\begin{ruledtabular}
\begin{tabular}{lcccc}
Graph & \(F\) & \(L_B\) & \(E\) & \(C\) \\
\hline
ER
& \(0.391\pm0.095\)
& \(0.110\pm0.041\)
& \(249\pm13\)
& \(0.070\pm0.014\) \\
SW
& \(0.529\pm0.038\)
& \(0.062\pm0.012\)
& \(240\pm0\)
& \(0.329\pm0.033\) \\
BA
& \(0.228\pm0.111\)
& \(0.229\pm0.135\)
& \(231\pm0\)
& \(0.169\pm0.035\) \\
RG
& \(0.344\pm0.174\)
& \(0.173\pm0.134\)
& \(240\pm19\)
& \(0.623\pm0.026\) \\
\end{tabular}
\end{ruledtabular}
\end{table}

\subsection{Effect of source--target graph distance}

The topology results suggest that graph structure affects the
forgetting--adaptation balance. To test whether this dependence can be reduced
to a simpler distance effect, we next controlled the minimum shortest-path
distance between the input nodes and the output node. For each Erdős--Rényi
graph, we selected output nodes satisfying
\[
d_{\mathrm{out}}
=
\min\{d(i_1,o),d(i_2,o)\},
\]
where \(i_1\) and \(i_2\) are the two input nodes and \(o\) is the output node.
We then repeated the sequential \(A\to B\) experiment for output nodes at
different values of \(d_{\mathrm{out}}\). Runs in which Task A was not learned
were excluded using the same criterion as above.

\begin{figure}[t]
\centering
\includegraphics[width=\columnwidth]{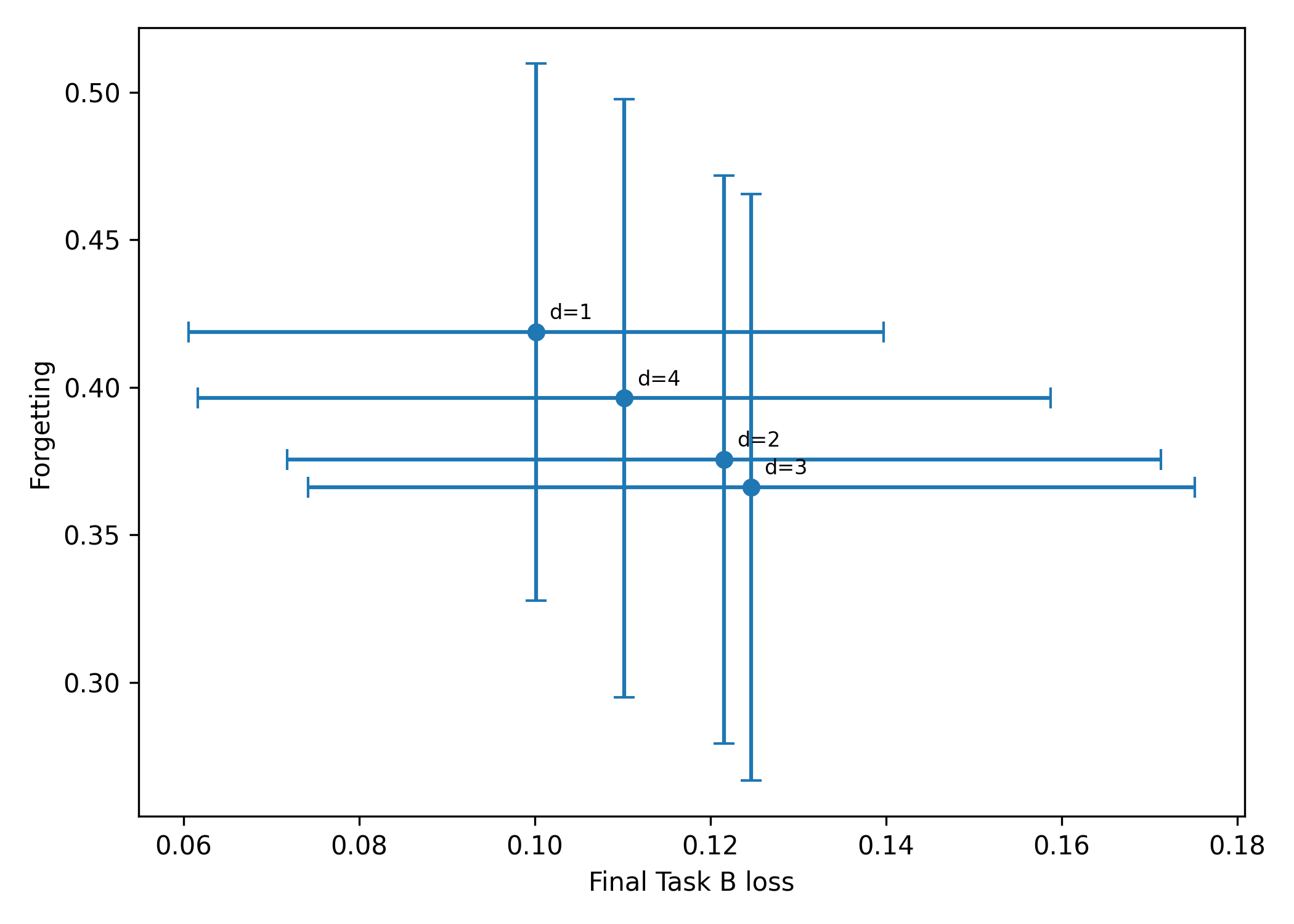}
\caption{
Effect of minimum input--output graph distance on the
forgetting--adaptation trade-off in Erdős--Rényi networks. Each point shows the
mean over successful Task-A runs, with error bars denoting one standard
deviation. The available distances show overlapping error bars, indicating that
shortest-path distance alone has only a weak effect on forgetting and final
Task B loss in this ensemble.
}
\label{fig:distance_tradeoff}
\end{figure}

The results are shown in Fig.~\ref{fig:distance_tradeoff}. Over the range of
distances available in the \(N=80\), \(p=0.08\) Erdős--Rényi graphs,
source--target distance has only a weak effect. Forgetting is largest for
directly connected or near-connected outputs, but the differences across
distances are comparable to the variability across graph realisations. Final
Task B loss also changes only modestly with distance. Thus, in this ensemble,
shortest-path distance alone does not account for the topology dependence
observed above. The broader graph structure, including degree heterogeneity,
clustering, and the availability of alternative conductive pathways, appears to
matter beyond the minimum input--output distance.

\begin{table}[t]
\caption{
Effect of minimum input--output graph distance in Erdős--Rényi networks. Values
are mean \(\pm\) standard deviation over successful Task-A runs.
}
\label{tab:distance}
\begin{ruledtabular}
\begin{tabular}{lccc}
\(d_{\mathrm{out}}\) & Runs & Forgetting & Task B loss \\
\hline
1 & 20 & \(0.419 \pm 0.091\) & \(0.100 \pm 0.040\) \\
2 & 20 & \(0.375 \pm 0.096\) & \(0.122 \pm 0.050\) \\
3 & 20 & \(0.366 \pm 0.099\) & \(0.125 \pm 0.050\) \\
4 & 14 & \(0.396 \pm 0.101\) & \(0.110 \pm 0.049\) \\
\end{tabular}
\end{ruledtabular}
\end{table}

\subsection{Effect of network size and training budget}

The preceding results show that forgetting must be interpreted jointly with
new-task adaptation. This is especially important when comparing networks of
different sizes, because a fixed optimisation budget may undertrain larger
systems. To separate size effects from optimisation-budget effects, we repeated
the \(A\to B\) experiment for Erdős--Rényi graphs with
\[
N\in\{40,80,160,320\},
\qquad
T\in\{300,1000,3000\},
\]
where \(T\) is the number of gradient steps used for each task. The expected
mean degree was kept approximately fixed by choosing \(p=6/(N-1)\).

\begin{figure}[t]
\centering
\includegraphics[width=\columnwidth]{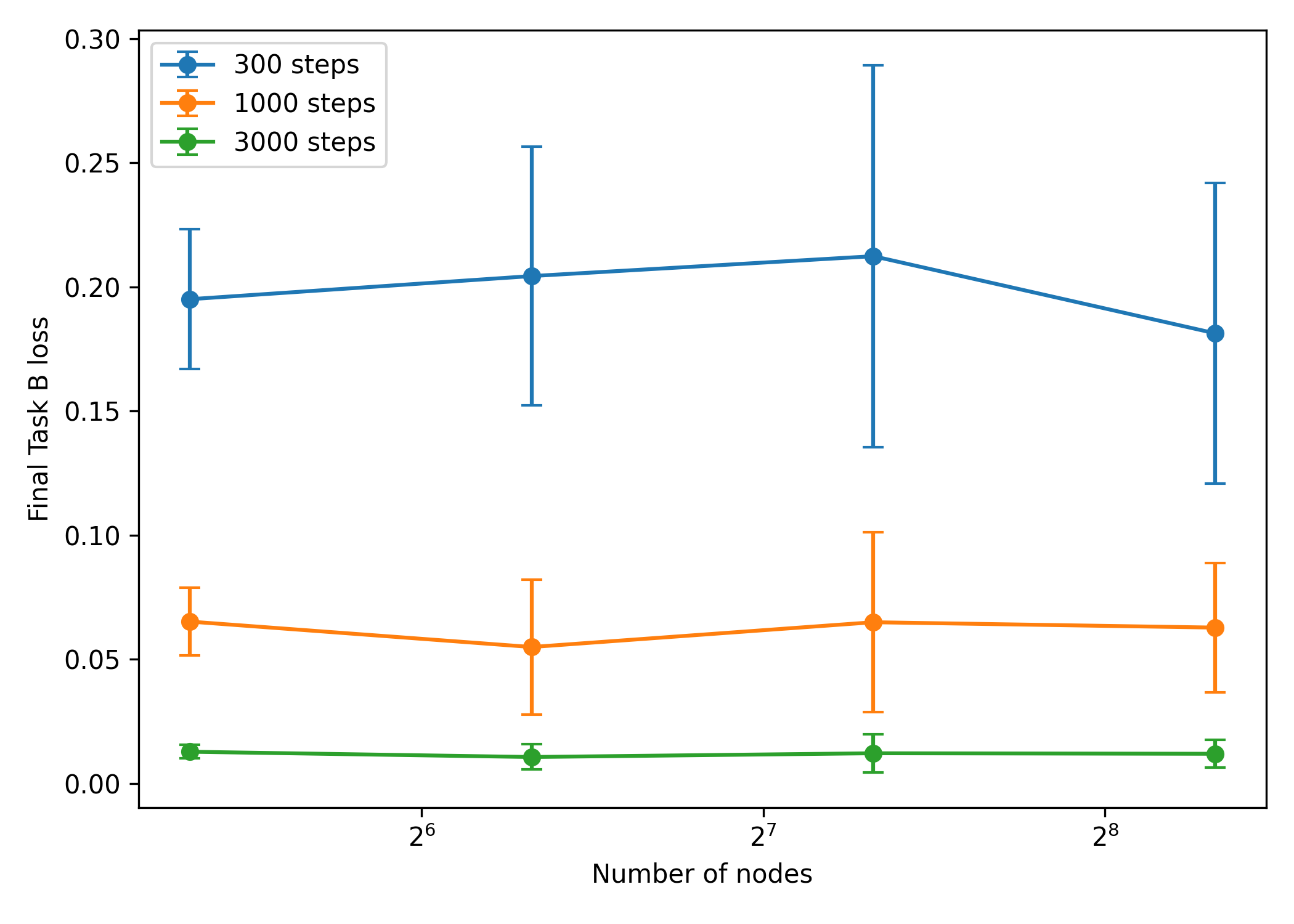}
\caption{
Final Task B loss as a function of network size for different training budgets.
Increasing the number of gradient steps strongly improves Task B learning,
whereas the dependence on \(N\) is weak over the range tested.
}
\label{fig:taskB_size_budget}
\end{figure}

\begin{figure}[t]
\centering
\includegraphics[width=\columnwidth]{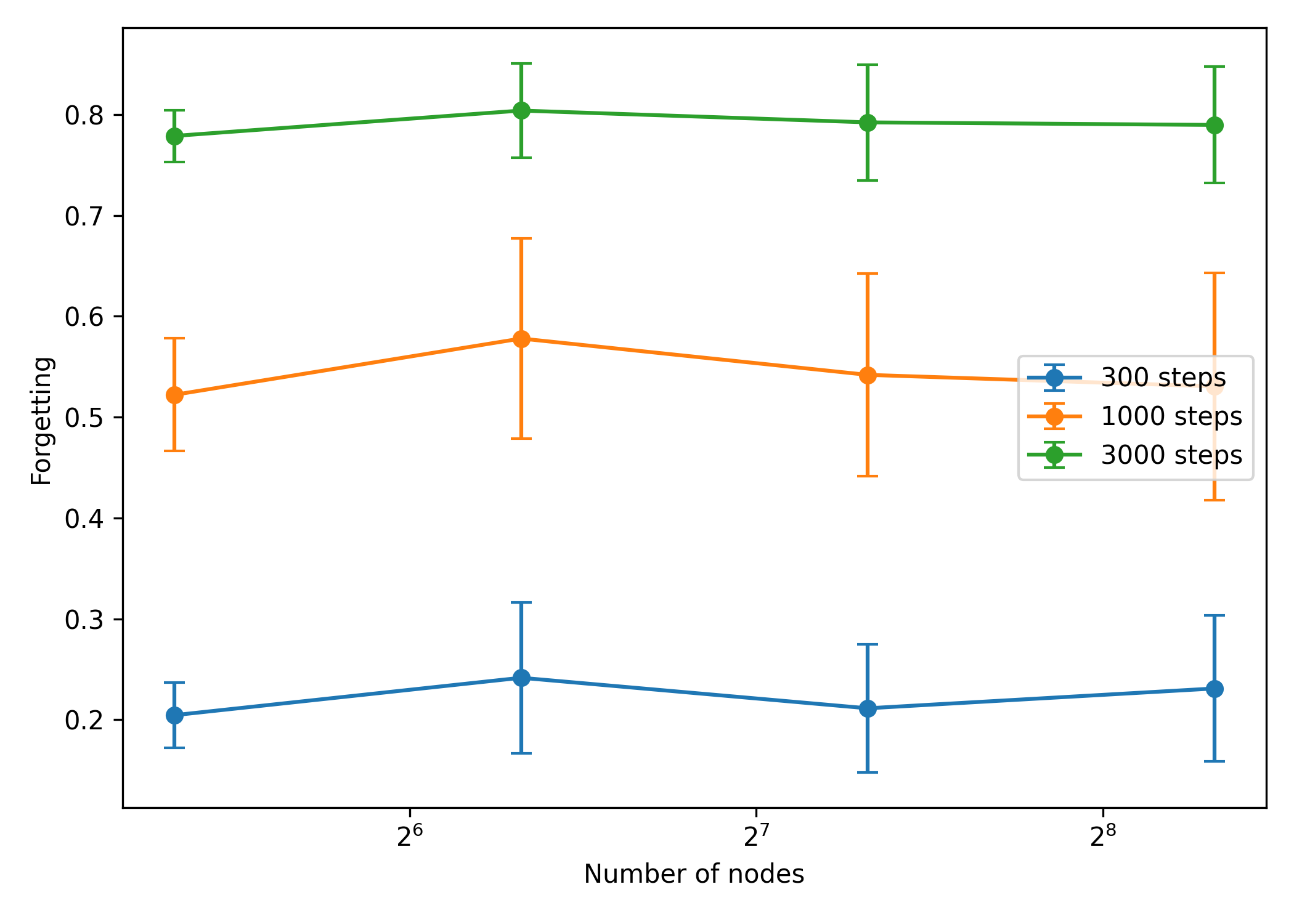}
\caption{
Forgetting as a function of network size for different training budgets.
Longer training on Task B produces substantially larger forgetting. The weak
dependence on \(N\), compared with the strong dependence on training budget,
indicates that fixed-budget size trends should be interpreted cautiously. For \(N=320,T=300\), one run was excluded because Task A was not learned under the criterion \(L_A^{\mathrm{before}}<0.25\).
}
\label{fig:forgetting_size_budget}
\end{figure}

The results are shown in Figs.~\ref{fig:taskB_size_budget}
and~\ref{fig:forgetting_size_budget}. Across this
range of sizes, network size has only a weak systematic effect compared with
the training budget. For \(T=300\), the final Task B loss remains relatively
large and forgetting is modest. Increasing the budget to \(T=1000\) improves
Task B learning but substantially increases forgetting. At \(T=3000\), Task B
is learned almost completely, but forgetting becomes severe across all network
sizes.

Thus, the apparent size dependence observed under a fixed training budget is
better interpreted as an optimisation-budget effect. Lower forgetting can arise
because the second task has not been learned fully. Once the training budget is
varied explicitly, the dominant trend is the same forgetting--adaptation
trade-off observed in the regularisation experiments.
\begin{table}[t]
\caption{
Effect of network size and training budget. Values are mean \(\pm\) standard
deviation across graph realisations, after excluding runs in which Task A was
not learned.
}
\label{tab:size_budget}
\begin{ruledtabular}
\begin{tabular}{cccc}
\(N\) & \(T\) & Forgetting & Task B loss \\
\hline
40 & 300 & \(0.205\pm0.033\) & \(0.195\pm0.028\) \\
40 & 1000 & \(0.522\pm0.056\) & \(0.065\pm0.014\) \\
40 & 3000 & \(0.779\pm0.026\) & \(0.013\pm0.003\) \\
80 & 300 & \(0.242\pm0.075\) & \(0.204\pm0.052\) \\
80 & 1000 & \(0.578\pm0.099\) & \(0.055\pm0.027\) \\
80 & 3000 & \(0.804\pm0.047\) & \(0.011\pm0.005\) \\
160 & 300 & \(0.211\pm0.064\) & \(0.212\pm0.077\) \\
160 & 1000 & \(0.542\pm0.100\) & \(0.065\pm0.036\) \\
160 & 3000 & \(0.792\pm0.057\) & \(0.012\pm0.008\) \\
320 & 300 & \(0.231\pm0.073\) & \(0.181\pm0.060\) \\
320 & 1000 & \(0.530\pm0.113\) & \(0.063\pm0.026\) \\
320 & 3000 & \(0.790\pm0.058\) & \(0.012\pm0.006\) \\
\end{tabular}
\end{ruledtabular}
\end{table}

\section{Discussion}

Taken together, the experiments show that catastrophic forgetting arises
naturally in differentiable resistor networks under sequential gradient-based
training, and that it admits a physically interpretable explanation. At the
coarsest level, training on Task B moves the conductance vector away from the
Task A solution. The task-similarity and random-task sweeps show that forgetting
is strongest when the second task reverses the output ordering imposed by the
first task. Gradient overlap provides a local parameter-space diagnostic of this
conflict: updates that reduce the new-task loss tend to oppose directions that
preserve the old task.

The random task ensemble provides a task-level interpretation of contradiction
in this minimal setting. Since Task A imposes the output ordering
\(y_1>y_2\), second tasks with the same ordering are cooperative, while tasks
with \(y_1<y_2\) are contradictory. The strong negative correlation between
forgetting and the Task-B target contrast \(c_B=y_1-y_2\) shows that forgetting
is controlled not merely by numerical target difference, but by whether the new
task reverses the functional ordering learned from the first task. Gradient
overlap remains useful as a parameter-space diagnostic, but the target contrast
is the clearer descriptor of task conflict in this two-input benchmark.

Regularisation partially suppresses forgetting, but only by limiting adaptation
to the new task. This is the stability--plasticity trade-off in its simplest
form. Uniform anchoring directly penalises movement away from the Task-A
solution, while gradient-weighted anchoring concentrates this penalty on
parameters assigned larger empirical importance weights. Once these weights are
normalised, the gradient-weighted method produces a pronounced trade-off curve:
it can suppress forgetting more strongly than uniform anchoring, but only by
moving further toward poor Task-B adaptation. Thus, the important point is not
that importance weighting eliminates forgetting, but that it moves the system to
a different region of the same forgetting--adaptation trade-off.

The parameter updates are not spread uniformly across the network. Instead,
most of the drift is concentrated in a relatively small subset of edges, and
those edges are disproportionately ones that already carry substantial current
under the Task-A solution. In physical terms, learning the second task rewires a
limited set of dominant conductive pathways. Forgetting therefore has a
structural interpretation: it reflects localised reconfiguration of the
transport backbone that originally implemented the earlier task.

The repeated-training experiment shows that forgetting is not irreversible.
After an \(A\rightarrow B\rightarrow A\) sequence, the network can recover much
of its original Task-A performance. This suggests that the system does not lose
representational capacity outright; rather, it is redirected toward the
currently trained task. At the same time, incomplete recovery under the same
optimisation budget indicates path dependence in parameter space: once
conductances have been reconfigured for a conflicting task, returning to the
earlier solution may require additional training effort.

The topology experiments show that graph structure changes the
forgetting--adaptation balance. Small-world graphs adapt strongly to the new
task but forget more, while scale-free graphs show lower forgetting with weaker
new-task learning. Random-geometric graphs display intermediate but more
variable behaviour. The source--target distance sweep shows that shortest-path
distance alone has only a weak effect in the Erdős--Rényi ensemble considered
here. Thus, topology matters, but its effect is not reducible to the minimum
input--output graph distance. Broader structural features such as clustering,
degree heterogeneity, and the availability of alternative conductive pathways
also shape the trade-off.

The size--budget sweep clarifies the interpretation of network-size effects.
At approximately fixed mean degree, increasing \(N\) from 40 to 320 produces
only weak systematic changes compared with the effect of training budget.
Longer Task-B training reduces the final Task-B loss but substantially
increases forgetting. This shows that lower forgetting under a fixed budget
should not be interpreted automatically as better memory retention; it may
instead reflect incomplete learning of the new task.

Overall, the revised experiments move the interpretation beyond the observation
that sequential training causes forgetting. In this benchmark, forgetting is
controlled by target-level task contradiction, by the degree of new-task
adaptation, and by graph-dependent physical reconfiguration. The resulting
picture is not a closed-form theory of forgetting, but it identifies the main
statistical and mechanistic factors controlling forgetting in differentiable
resistor networks.
\section{Conclusion}

We have studied sequential learning in differentiable resistor networks and
shown that even this minimal graph-Laplacian physical system exhibits
catastrophic forgetting under ordinary gradient-based training. The forgetting
is physically interpretable: sequential learning produces localised changes in
conductance parameters, and the largest changes occur preferentially on
high-current edges that form dominant transport pathways.

Regularisation reduces forgetting, but only through a stability--plasticity
trade-off. Uniform anchoring and normalised gradient-weighted anchoring both
move the system along a forgetting--adaptation curve: stronger regularisation
protects the old task but worsens learning of the new one. The random task
ensemble shows that contradiction is controlled by target ordering: second tasks
that reverse the ordering imposed by the first task produce the strongest
forgetting. The topology experiments show that the trade-off also depends on graph
structure. Small-world, scale-free, Erdős--Rényi, and random-geometric networks
occupy different regions of the forgetting--adaptation plane, showing that
network architecture affects both memory retention and new-task acquisition.
The size--budget sweep further shows that training budget is a dominant factor:
more complete learning of the second task produces stronger forgetting across
all tested network sizes.

These results position differentiable resistor networks as compact and
physically interpretable testbeds for continual learning in tunable matter.
Future work should extend the framework to longer task sequences, richer task
families, controlled source--target graph distances, larger networks with
matched convergence criteria, and locality-constrained learning rules closer to
physical hardware implementation.

\begin{acknowledgments}
The author acknowledges support from the Mathematics Applications Consortium for Science and Industry (MACSI), Department of Mathematics and Statistics, University of Limerick.
\end{acknowledgments}

\section*{Data availability}
The code and generated data supporting the findings of this study are publicly
available in the Zenodo-archived repository \emph{Physical Learning in Resistor
Networks}, version 1.1.0~\cite{ibrahim_resistor_code_2026}. The corresponding GitHub repository is available at https://github.com/Manirmaths/physical-learning-resistor-networks.

\bibliographystyle{apsrev4-2}
\bibliography{references}

\end{document}